\documentclass[runningheads]{llncs}
\usepackage[T1]{fontenc}
\usepackage{cite}
\usepackage{graphicx}
\usepackage{subfigure}
\usepackage{colortbl}
\usepackage{graphicx}
\usepackage{float}
\usepackage{subfigure}
\usepackage{amssymb}
\usepackage{multicol}
\usepackage{multirow}
\usepackage{setspace}
\usepackage{marvosym}
\usepackage[marginal]{footmisc}
\usepackage{url}

\begin{document}
\mainmatter              
\title{S$^2$R: Exploring a Double-Win Transformer-Based Framework for Ideal and Blind Super-Resolution}
\titlerunning{S$^2$R}  
%
\author{Minghao She \and Wendong Mao\textsuperscript{(\Letter)} \and
Huihong Shi \and Zhongfeng Wang\textsuperscript{(\Letter)}, \emph{Fellow, IEEE}}
\authorrunning{Minghao She} 
%
\tocauthor{Minghao She, Wendong Mao, Huihong Shi, Zhongfeng Wang}
%
\institute{School of Electronic Science and Engineering, Nanjing University, P.R. China\\
\email{mhshe@smail.nju.edu.cn, wdmao@smail.nju.edu.cn, shihh@smail.nju.edu.cn, zfwang@nju.edu.cn}}

\maketitle             
\begin{abstract}
Nowadays, deep learning based methods have demonstrated impressive performance on ideal super-resolution (SR) datasets, but most of these methods incur dramatically performance drops when directly applied in real-world SR reconstruction tasks with unpredictable blur kernels. To tackle this issue, blind SR methods are proposed to improve the visual results on  random blur kernels, which causes unsatisfactory reconstruction effects on ideal low-resolution images similarly. In this paper, we propose a double-win framework for ideal and blind SR task, named S$^2$R, including a light-weight transformer-based SR model (S$^2$R transformer) and a novel coarse-to-fine training strategy, which can achieve excellent visual results on both ideal and random fuzzy conditions. On algorithm level, S$^2$R transformer smartly combines some efficient and light-weight blocks to enhance the representation ability of extracted features with relatively low number of parameters. For training strategy, a coarse-level learning process is firstly  performed to improve the generalization of the network with the help of a large-scale external dataset, and then, a fast fine-tune process is developed to transfer the pre-trained model to real-world SR tasks by mining the internal features of the image. Experimental results show that the proposed S$^2$R outperforms other single-image SR models in ideal SR condition with only 578K parameters. Meanwhile, it can achieve better visual results than regular blind SR models in blind fuzzy conditions with only 10 gradient updates, which improve convergence speed by 300 times, significantly accelerating the transfer-learning process in real-world situations. Codes will be found in https://github.com/berumotto-vermouth/S2R.\footnote{This work was supported in part by National Key R\&D Program of China under Grant 2022YFB4400604.}\footnote{Corresponding author: Wendong Mao and Zhongfeng Wang}
\end{abstract}
%
\section{Introduction}
Super-Resolution (SR) is proposed to increase the resolution of low-quality images and enhance their clarity. As a fundamental low-level vision task, single image super-resolution (SISR), which aims to recover plausible high-resolution (HR) images from their counterpart low-resolution (LR) images, has attracted increasing attention. With the remarkable success of convolutional neural networks (CNNs), various deep learning based methods with different network architectures and training strategies have been proposed for SISR and achieved prominent visual results. Most of them are optimized upon a large number of external training dataset, which uses the fixed bicubic operation to downsample HR images for obtaining LR images and constructing paired training datasets. For imitating low-resolution images in ideal conditions, in regular tasks, the LR images are generated by down-sampling in the noise-free bicubic condition. In this way, several previous deep learning based methods~\cite{liang2021swinir,zhang2022efficient,dong2016accelerating,shi2016real,lai2017deep} show excellent performance in ideal conditions.

In real-world situations, noise and blur are inevitable and influenced by actual equipment, such as camera lens condition, shooting conditions, which is sometimes different from the ideal one. And  these blur kernels affected by real-world conditions are unpredictable in most cases. Thus, to imitate the actual degradation blur condition of real-world images, most blind SR methods~\cite{begin2004blind,he2011single,wang2005patch} assume that the LR input image is obtained by down-sampling the counterpart HR image by an unpredictable blur kernel. As revealed in ~\cite{efrat2013accurate}, learning-based methods will suffer severe performance drops when the blur kernels in the test phase and the training phase are inconsistent. This kind of kernel mismatch will introduce undesired artifacts to output images. Thus, the problem with unpredictable blur kernels, also known as blind SR, has limited the application of some deep learning based SR methods in real-world situations. To solve this limitation, several novel methods have been proposed. For instance, some blind SR methods~\cite{begin2004blind,he2011single,wang2005patch} are model-based, which introduce prior knowledge to the deep learning area and usually involve complicated optimization procedures. These methods do not calculate the SR kernel directly, but assume that SR networks are robust and transferable to variations in the downsampling kernels.~\cite{michaeli2013nonparametric} exploited the recurrence of small image patches across scales of a single image to estimate the unknown SR kernel directly from the LR input image. However, it fails when the SR scale is larger than 2, and the runtime is very long. ZSSR~\cite{shocher2018zero} does not adopt the regular training process on the external datasets and trains from scratch for each test image, so that specific models pay more attention to information in the picture. Nevertheless, the runtime is greatly increased, limiting its application in the real scene. The current challenge is that in the above SR models, no one can show ideal performance simultaneously on both SISR and blind SR tasks. Thus, it's urgent to propose an SR method to solve this challenge.

In this paper, we propose a novel framework S$^2$R. In it, inspired by the great success of transformer in computer vision tasks, a light-weight transformer-based model is proposed to achieve excellent performance in the ideal LR condition, and a novel coarse-to-fine training strategy is performed to boost the scalability for guaranteeing the real-world applications and achieving fast convergence by extremely few gradient updates. The detailed contributions are shown as follow:
\begin{itemize}
    \item[\textbullet] To achieve excellent performance in ideal and blind SR conditions, we propose a new framework S$^2$R, containing a light-weight transformer-based model where the excellent visual results can be achieved in ideal conditions, and a novel coarse-to-fine training strategy where the proposed model can be extended to the real-world applications. 
    \item[\textbullet] For the network architecture, we propose a light-weight transformer-based model for super resolution, which enhances the representational power by combining several light-weight efficient blocks, achieving good performance in ideal SISR tasks.
    \item[\textbullet] To boost the scalability of the proposed model in real-world situations, we further propose a new coarse-to-fine training strategy, which is to pre-train the proposed model on a large-scale external dataset, and then, perform transfer-learning on a new internal real-world dataset to take advantage of external and internal learning, realizing better performance and fast-adaption for real-world applications. 
    \item[\textbullet] The experiment result shows the superiority of the proposed framework. For SISR tasks, compared to sort-of-the-art (SOTA) transformer-based models, our methods can achieve comparable performance with minimum number of parameters parameters. And for blind SR tasks, our methods can achieve better performance and clearer outputs than other blind SR models. Moreover, the proposed framework reduces the number of backpropagation gradient updates by 300 times.
\end{itemize}
\section{Related Work}
\subsection{Single Image Super Resolution}
SISR is based on the image degradation model as
\begin{equation}
    I_{LR} = (I_{HR} \ast k_s)\downarrow_s,
\end{equation}
where I$_{HR}$, I$_{LR}$, k$_s$, ${s}$ denote HR, LR image, SR kernel, and scaling factor, respectively. Recently, CNN-based methods~\cite{dong2016accelerating,shi2016real,lai2017deep,kim2016deeply,tai2017image} have demonstrated impressive performance in SR tasks. SRGAN~\cite{ledig2017photo} first introduced residual blocks into SR networks. EDSR~\cite{lim2017enhanced} also used a deep residual network for training SR model but removed the unnecessary batch normalization layer in the residual block. Zhang \emph{et al.}~\cite{zhang2018image} achieved better performance than EDSR by introducing the channel attention to residual block to form a very deep network. Haris \emph{et al.}~\cite{haris2018deep} proposed deep back-projection networks to exploit iterative up and down sampling layers, providing an error feedback mechanism for projection errors at each stage. SRMD~\cite{zhang2018learning} proposed a stretching strategy to integrate kernel and noise information to cope with multi-degradation kernels in an SR network. The breakthrough of transformer networks in natural language processing (NLP) inspires researchers to use self-attention (SA) in vision tasks. Thus, the transformer-based SR models come into being. SwinIR~\cite{liang2021swinir} adapted the Swin Transformer~\cite{liu2021swin} to image restoration, which combines the advantage of both CNNs and transformers. ELAN~\cite{zhang2022efficient} based on SwinIR~\cite{liang2021swinir}, removed redundant designs such as masking strategy and relative position encoding to make the network more slim. Reducing the number of parameters of transformer-based methods and alleviating performance drops is a feasible direction. Based on this, we introduce some light-weight blocks to reduce parameters further, and boost the representational power, achieving good visual results.
\subsection{Blind Super-Resolution}
Compared to SISR, blind SR assumes that the degradation blur kernels are unpredictable. In real-world applications, the SR kernel of images is influenced by the sensor optics. In recent years, some blind SR methods~\cite{wang2005patch,he2011single} introduce prior knowledge into the deep learning area. However, these methods are to make their models more robust to variations, rather than explicitly calculating the SR kernel. In contrast, ~\cite{michaeli2013nonparametric} estimates the kernel based on the recurrence of small image patches (5$\times$5, 7$\times$7) across scales of a single image but fails for SR scale factors larger than 2. KernelGAN~\cite{bell2019blind} based on Interal-GAN~\cite{shocher2019ingan} estimates the SR kernel that best preserves the distribution of patches across scales of the LR image. ZSSR~\cite{shocher2018zero} trains a small full convolution network for each image from scratch to learn image-specific internal structure, rather than adapting the training process to big data. However, it drastically increases the runtime for thousands of backpropagation gradient updates at test time. In contrast, the proposed methods can significantly reduce the runtime with the help of our pre-train model on a large-scale external dataset. 
\section{The Proposed S$^2$R Framework}
\begin{figure}
    \centering
    \subfigure[The proposed S$^2$R transformer]{
        \begin{minipage}[t]{0.9\linewidth}
        \centering
        \includegraphics[width=0.9\linewidth]{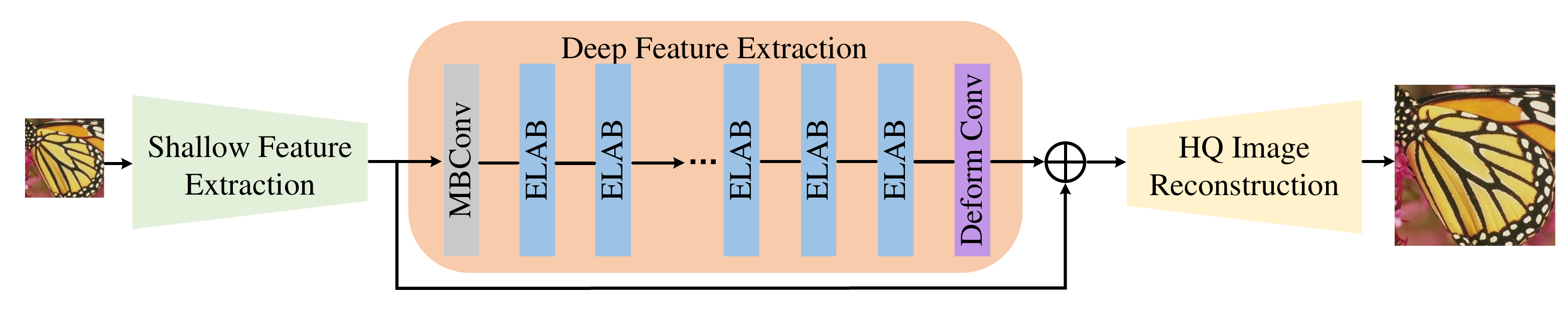}
        \label{architecture}
        \end{minipage}%
        }%
        
    \subfigure[The proposed coarse-to-fine training strategy]{
        \begin{minipage}[t]{0.9\linewidth}
        \centering
        \includegraphics[width=0.9\linewidth]{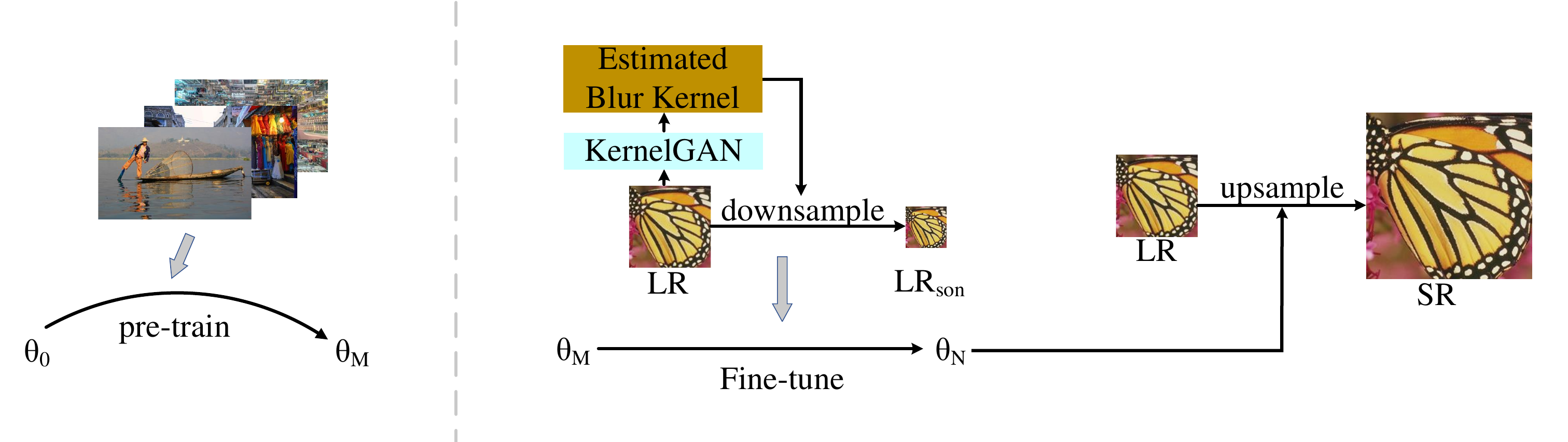}
        \label{training_recipe}
        \end{minipage}%
        }%

    \caption{Illustration of the proposed S$^2$R framework. (a) The architecture of the proposed model, S$^2$R transformer. (b) The overall pipeline of the proposed coarse-to-fine training strategy.}
    \label{framework}
\end{figure}

In this section, we introduce a double-win framework for ideal and blind SR (S$^2$R), which consists of a light-weight transformer-based model (S$^2$R transformer) and a novel coarse-to-fine training strategy, as shown in Fig. \ref{framework}. With the proposed training strategy, the proposed model can achieve significant performance with a few backpropagation gradient updates and great scalability for the real-world applications. The learning process of the framework is mainly divided to two stages, coarse-level learning on the general dataset and fast fine-tune on real-world images. The first stage is to pre-train the proposed model on a large scale external dataset to guarantee its great visual results on ideal LR conditions and fast-adaption when the proposed model is extended to the real-world situations. The second stage is to boost the scalability of our model, to extend the real-world applications.
\subsection{Network Architecture}
As shown in Fig. \ref{architecture}, the proposed S$^2$R transformer contains three modules, a shallow feature extraction module (SF), a deep feature extraction module (DF) and a high-quality image reconstruction module (RC). Specifically, given a low-resolution (LR) input I$_{LR}\in \mathbb{R^{3{\times}H{\times}W}}$, where H and W are the height and width of the LR image, respectively, we first use the shallow feature extraction module denoted by H$_{SF}(\cdot)$, which only consists of a single 3$\times$3 convolution, to extract local feature I$_{local} \in \mathbb{R^{C{\times}H{\times}W}}$, the deep feature extraction module consists of a mobile bottleneck convolution (\textbf{MBConv}), denoted by H$_{mbconv}$, N cascaded efficient long-range attention blocks (\textbf{ELAB}), denoted by H$_{ELAB}$, a deformable convolution~\cite{dai2017deformable} (\textbf{Deform Conv}), denoted by H$_{deform}$ and a 3$\times$3 convolution:
\begin{equation}
     I_{local} = H_{SF}(I_{LR}), 
\end{equation}
where C is the feature channel number.

I$_{local}$ then goes to the deep feature extraction module, denoted by H$_{DF}(\cdot)$. That is:
\begin{equation}
    I_{deep} = H_{DF}(I_{local}),
\end{equation}
where I$_{deep} \in \mathbb{R^{C{\times}H{\times}W}}$ denotes the output of the deep feature extraction module. More specifically, intermediate features I$_{mbconv}$, I$_{deform}$, I$_1$, I$_2$, $\dots$, I$_N$, and the output deep feature I$_{deep}$ are extracted block by block as
\begin{equation}
    \centering
    \begin{array}{rcl}
        I_{mbconv} &=& H_{mbconv}(I_{local}) \\
        I_i &=& H_{ELAB_i}(I_{i-1}),\qquad i=1,2, \dots, N \\
        I_{deform} &=& H_{deform}(I_N) \\
        I_{deep} &=& H_{CONV}(I_{deform}),
    \end{array}
\end{equation}
where H$_{ELAB_i}(\cdot)$ denotes the \emph{i}-th ELAB and H$_{CONV}$ is the last convolutional layer. Finally, taking I$_{deep}$ and I$_{local}$ as inputs, the HR image I$_{HQ}$ is reconstructed as:
\begin{equation}
    I_{RHQ} = H_{RC}(I_{deep} + I_{local}),
\end{equation}
where H$_{RC}$ is the reconstruction module. Here we choose a sub-pixel convolution layer~\cite{shi2016real} to upsample the feature. The proposed model can be optimized with the commonly used loss functions for SR, such as L$_2$~\cite{dong2015image} and L$_1$~\cite{lai2017deep, lim2017enhanced, zhang2018residual}. For simplicity, we choose L$_1$ as our loss function by minimizing the L$_1$ pixel loss.
\subsubsection{ELAB}
Despite the great success of transformer on computer vision tasks, it greatly suffers from explosive parameters and high computational complexity. Thus, inspired by~\cite{zhang2022efficient}, we introduce ELAB to achieve a good performance in the case of a relatively low number of parameters. In ELAB, the regular multi-layer perception is replaced with two shift-conv~\cite{wu2018shift} with a simple ReLU activation. It can help enlarge the receptive field of the proposed model preliminarily, while sharing the same arithmetic complexity as two cascaded 1$\times$1 convolutions. Moreover, to significantly reduce parameters, the accelerated attention mechanism (ASA) is utilized. In the self-attention calculation procedure, three independent 1$\times$1 convolutions, denoted by $\theta, \phi, \psi$, are employed to map the input feature X into three different feature maps. In ELAB, $\theta$ is set the same as $\phi$, which can save 1$\times$1 convolution in each self-attention. For compensating for performance drops caused by the reduced number of parameters, instead of the regular self-attention, the group-wise multi-scale self-attention (GMSA) is used. The computational complexity of the regular window-based self-attention~\cite{liu2021swin} is determined by the window size M. Thus, in GMSA, the input feature map is divided into three groups, and then set a flexible window size for each group. In this way, the relative position bias in regular transformer-based SISR models, which makes models fragile to the resolution change~\cite{xie2021segformer}, can be removed. Based on this, the proposed model can also extract information with different scales and be more flexible to the input, which makes it feasible to extend the proposed model to real-world applications. Finally, batch normalization (BN)~\cite{ioffe2015batch} is utilized to replace layer normalization (LN), accelerating the calculation by avoiding fragmenting the calculation into many element-wise inefficient operations.
\subsubsection{MBConv}
For further performance improvement, inspired by the intuition that the bottleneck actually contains all the necessary information, MBConv is introduced to enhance the representational power. It is based on depthwise separable convolution~\cite{chollet2017xception}, which inherently has fewer parameters than the full convolution. Thus, it will not bring lots of parameters. Furthermore, a squeeze-and-excitation optimization~\cite{hu2018squeeze} is added to enlarge the receptive field further. For its inverted design, it is more memory efficient which will further decrease running time and parameter count for bottleneck convolution. Therefore, this light-weight block can help the proposed model get a larger receptive field and better performance. Recent studies~\cite{graham2021levit,xiao2021early} suggest that using convolutions in early stages benefits the performance of Vision Transformers. Thus, we follow this design to put MBConv in early stages.
\subsubsection{Deform Conv}
To extract irregular structural features, a Deform Conv is added. It is based on the idea of augmenting the spatial sampling in the modules with additional offsets and learning the offsets from the target tasks, without introducing additional supervision. Thus, it is lightweight on its own, without sacrificing computing resources to trade off the performance. Furthermore, as the receptive field is further enlarged, it can facilitate generalization to new tasks possessing unpredictable geometric transformations, improving the performance in SISR tasks, and supporting subsequently migrating the proposed model into the real-world applications.
\subsection{The Proposed Coarse-to-Fine Training Strategy}
To guarantee impressive performance in ideal SR conditions while extending the proposed model to applications in the real world, we propose a coarse-to-fine training strategy. Fig. \ref{training_recipe} shows the pipeline of the proposed fast training strategy. It consists of two stages, coarse-level learning on the general dataset and fast fine-tune on real-world images. 

\vspace{-1.5em}
\subsubsection{Coarse-Level Learning on the General Dataset}
Considering that most blind SR methods directly train on the input test dataset, without fully exploiting the large-scale external dataset. Thus, as shown on the left side of the dotted line in Fig. \ref{training_recipe}, we firstly pre-train S$^2$R transformer on the external dataset (noise-free ``bicubic" dataset). This procedure can guarantee the good performance on SISR tasks, even if the proposed model is migrated to the real-world applications.

\subsubsection{Fast Fine-Tune on Real-World Images}
 Avoiding deviating from the goal of calculating the blur kernel to deal with blind SR tasks, rather than just make the proposed model robust to variations in the degradation LR kernel, it's necessary to estimate the SR kernel that best preserves the distribution of patches across scales of the LR image, which can compensate for the performance drop caused by the blur kernel mismatch. Recent studies~\cite{bell2019blind, michaeli2013nonparametric, zontak2011internal} have exploited the kernel estimation well, we estimate the blur kernel by the method in ~\cite{bell2019blind}, where its \emph{Generator} is trained to generate a downscaled image of the LR test image, such that its \emph{Discriminator} cannot distinguish between the patch distribution of the downscaled image and the origin. 

The diversity of the LR-HR relations within a single image is  significantly small, related to noise and blur kernel. Thus, as shown on the right side of the dotted line in Fig. \ref{training_recipe}, given an LR real-world test image I$_{LR}$, we use the estimated blur kernel ${K}$ to generate the lower resolution image I$_{LRson}$, and synthesize them as a data pair LR-LR$_{son}$. Then, we use the pre-trained model generated in \emph{Coarse-Level Learning on the General Dataset} to generate the corresponding high-resolution version I$_{SR}$ of the I$_{LRson}$, and optimize to learn the residual between I$_{SR}$ and I$_{LR}$. With the help of training with the external dataset and the proposed light-weight transformer-based model, the runtime can be drastically reduced. Thus, only 10 iterations are needed. Furthermore, a learning-rate update policy is performed, where we start with a learning rate of $\beta_0$, and then, change to $\beta_1$ when the iteration is not larger than 4, and keep $\beta_2$ until the end. Finally, the method in~\cite{lim2017enhanced}, where 8 different outputs for several rotations and flips of the test image I are generated and combined, is utilized to improve the performance on images in real-world situations. Moreover, their mean is replaced with the median of these 8 outputs. We further combine it with the back-projection of ~\cite{irani1991improving}, so that each of the 8 output images goes through multiple iterations of the back-projection and the final correction of the median image can also be done through the back projection.

Compared to conventional blind SR methods, such as ZSSR~\cite{shocher2018zero}, and conventional SISR methods, such as SwinIR~\cite{liang2021swinir}, the advantages of the proposed framework are twofold:
\begin{itemize}
    \item [\textbullet] In terms of network architecture, compared to ZSSR, whose backbone is a fully convolutional network, our model is based on transformer blocks which have more powerful learning and representation, thus achieving  better performance than ZSSR.
    \item[\textbullet] In terms of training strategy, ZSSR trains from scratch on the small dataset for implementing the SR task, while we pre-train the proposed model on the external dataset, and then, transfer-learning is performed on the proposed model to extend our model to a new SR task. It makes our model fast-adaption on downstream tasks and real-world applications and further boosts the performance of the proposed model.
\end{itemize} 
\section{Experiments $\&$ Results}
\subsection{Training Details}
For the pre-trained stage, we employ the DIV2K dataset~\cite{timofte2017ntire} with 800 training images to train our model, which consists of 10 ELABs with 60 channels, an MBConv, and a Deformable Conv. The model is trained using the ADAM~\cite{kingma2014adam} optimizer with the learning rate $\beta$=2$\times$10$^{-4}$. Following ELAN~\cite{zhang2022efficient}, our multi-scale window sizes are set: 4$\times$4, 8$\times$8 and 16$\times$16. The model training is conducted by Pytorch on NVIDIA 2080Ti GPUs. For the transfer-learning stage, our model is fine-tuned for 10 iterations, using the ADAM~\cite{kingma2014adam} optimizer with the learning rate $\beta_0$=2$\times$10$^{-2}$, and then, change to $\beta_1$=1$\times$10$^{-2}$ when the iteration is not larger than 4, and keep $\beta_2$=5$\times$10$^{-3}$ until the end.
\subsection{Evaluations on Ideal Super-Resolution Dataset}
We evaluate our model with several popular SISR models, including CNN-based models CARN~\cite{ahn2018fast}, IMDN~\cite{hui2019lightweight}, LAPAR-A~\cite{li2020lapar}, LatticeNet~\cite{luo2020latticenet}, and transformer-based models SwinIR-light~\cite{liang2021swinir} and ELAN-light~\cite{zhang2022efficient} on popular benchmarks: Set5~\cite{bevilacqua2012low}, Set14~\cite{zeyde2012single}, BSD100~\cite{martin2001database}, Urban100\cite{huang2015single} and Manga109\cite{matsui2017sketch}, for performance comparison. Peak Signal-to-Noise Ratio (PSNR) and Structural SIMilarity (SSIM) are used as evaluation metrics, which are calculated on the Y channel after converting RGB to YCbCr format.
\subsubsection{Quantitative comparison}
The quantitative indexes of different methods are reported in Table \ref{tbl:SISR_tasks}. The transformer-based methods outperform many CNN-based methods, by exploiting the self-similarity of images. However, among them, SwinIR-light~\cite{liang2021swinir} suffers from its high number of parameters, which will place a heavy burden on the deployment, limiting its applications in real-world situations. Compared to it, ELAN-light~\cite{zhang2022efficient} improves the performance and accelerates the inference for real-world applications. Among all light-weight CNN-based SR models, parameters of the proposed model are a little more than LAPAR-A~\cite{li2020lapar}, while the PSNR/SSIM of the proposed model outperforms all of them greatly. Among all light-weight transformer-based SR models, our model has the lowest number of parameters 578K. Compared to the SOTA model ELAN~\cite{zhang2022efficient}, the PSNR/SSIM of the proposed model is comparable to it with almost identical visuals as shown in Fig. \ref{fig:ELAN} and Fig. \ref{fig:SIBSR}. It's worth mentioning that our model outperforms ELAN significantly on real-world tasks, detailed can be seen in Section \ref{4.3}.
\begin{table}[ht]
\centering
\caption{Performance comparison of different light-weight SR models in ideal SR conditions. \#Params indicates the total number of network parameters. The efficiency proxies (\#Params) is measured under the setting of upscaling SR images to 1280$\times$720 resolution. Best and second-best PSNR/SSIM indexes are marked in {\color{red}red} and {\color{blue}blue} colors, respectively.}
\resizebox{\linewidth}{!}{
\begin{tabular}{|c|c|c|c|c|c|c|c|}
\hline
Scale & Model        & \begin{tabular}[c]{@{}c@{}}\#Params\\      (K)\end{tabular} & \begin{tabular}[c]{@{}c@{}}Set5~\cite{bevilacqua2012low}\\      PSNR/SSIM\end{tabular} & \begin{tabular}[c]{@{}c@{}}Set14~\cite{zeyde2012single}\\      PSNR/SSIM\end{tabular} & \begin{tabular}[c]{@{}c@{}}BSD100~\cite{martin2001database}\\      PSNR/SSIM\end{tabular} & \begin{tabular}[c]{@{}c@{}}Urban100\cite{huang2015single}\\      PSNR/SSIM\end{tabular} & \begin{tabular}[c]{@{}c@{}}Manga109\cite{matsui2017sketch}\\      PSNR/SSIM\end{tabular} \\ \hline
$\times$2    & CARN~\cite{ahn2018fast}         & 1592                                                        & 37.76/0.9590                                                  & 33.52/0.9166                                                   & 32.09/0.8978                                                    & 31.92/0.9256                                                      & 38.36/0.9765                                                      \\
$\times$2    & IMDN~\cite{hui2019lightweight}         & 694                                                         & 38.00/0.9605                                                  & 33.63/0.9177                                                   & 32.19/0.8996                                                    & 32.17/0.9283                                                      & 38.88/0.9774                                                      \\
$\times$2    & LAPAR-A~\cite{li2020lapar}      & 548                                                         & 38.01/0.9605                                                  & 33.62/0.9183                                                   & 32.19/0.8999                                                    & 32.10/0.9283                                                      & 38.67/0.9772                                                      \\
$\times$2    & LatticeNet~\cite{luo2020latticenet}   & 756                                                         & 38.06/0.9607                                                  & 33.70/0.9187                                                   & 32.20/0.8999                                                    & 32.25/0.9288                                                      & \mbox{—} / \mbox{—}                                                             \\
$\times$2    & SwinIR-light~\cite{liang2021swinir} & 878                                                         & 38.14/0.9611                                                  & 33.86/0.9206                                                   & \color{red}{32.31}/\color{blue}{0.9012}                                                    & \color{red}{32.76/0.9340}                                                      & \color{red}{39.12/0.9783}                                                      \\
$\times$2    & ELAN-light~\cite{zhang2022efficient}   & 582                                                         & \color{red}{38.17/0.9611}                                                  & \color{red}{33.94/0.9207}                                                   & 32.30/\color{blue}{0.9012}                                                    & \color{red}{32.76/0.9340}                                                      & \color{blue}{39.11/0.9782}                                                      \\
$\times$2    & S$^2$R transformer (Ours)         & 578                                                      & \color{blue}{38.15/0.9611}                                                  & \color{blue}{33.88/0.9206}                                                   & \color{blue}{32.30}/\color{red}{0.9013}                                                    & \color{blue}{32.61/0.9333}                                                                  & 39.01/0.9780  \\ \hline                                                               
\end{tabular}
}
\label{tbl:SISR_tasks}
\end{table}
\subsubsection{Qualitative comparison}
\begin{figure}[h]
    \centering
    \begin{minipage}[b]{0.29\textwidth}
    \centering
            \subfigure[Barbara]{
            \includegraphics[width=\textwidth]{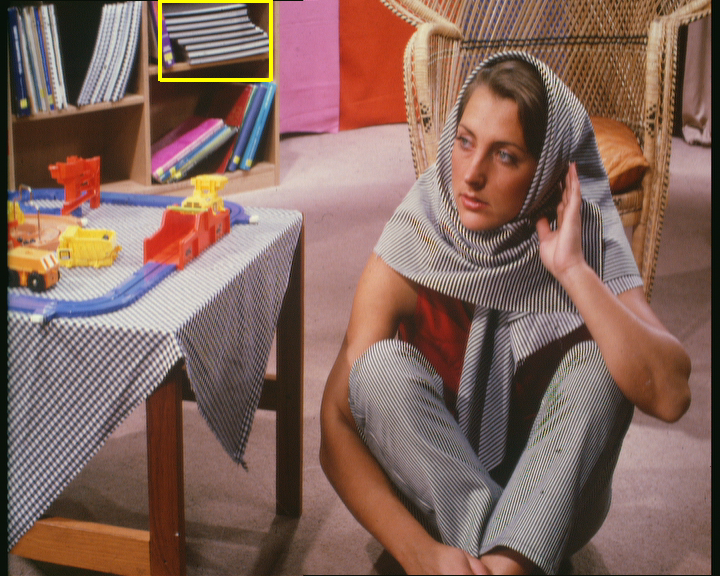}
        }
    \end{minipage}
    \begin{minipage}[b]{0.7\textwidth}
    \centering
            \subfigure[CARN]{
            \includegraphics[width=0.23\textwidth]{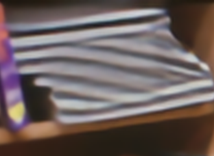}
            \label{fig:CARN}
        }
        \subfigure[IMDN]{
            \includegraphics[width=0.23\textwidth]{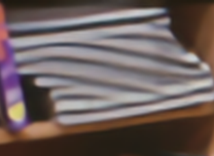}
            \label{fig:IMDN}
        }
        \subfigure[SwinIR-light]{
            \includegraphics[width=0.23\textwidth]{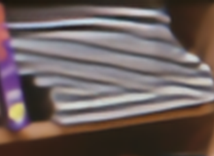}
            \label{fig:SwinIR}
        }\\
        \subfigure[ELAN-light]{
            \includegraphics[width=0.23\textwidth]{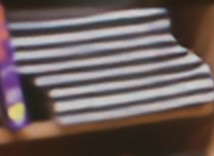}
            \label{fig:ELAN}
        }
        \subfigure[S$^2$R (Ours)]{
            \includegraphics[width=0.23\textwidth]{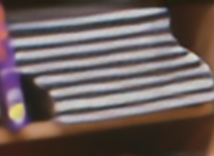}
            \label{fig:SIBSR}
        }
        \subfigure[GT]{
            \includegraphics[width=0.23\textwidth]{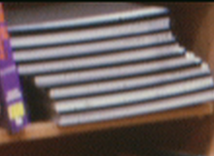}
            \label{fig:HR}
        }
    \end{minipage}
\label{crop_barbara}
\caption{Qualitative comparison of state-of-the-art light-weight SR models for $\times$4 upscaling in ideal super-resolution conditions, where the enlarged local patch of output image is shown in the right. (b) The result of CARN~\cite{ahn2018fast}. (c) The result of IMDN~\cite{hui2019lightweight}. (d)The result of SwinIR-light~\cite{liang2021swinir}. (e) The results of ELAN-light~\cite{zhang2022efficient}. (f) The result of ours. (g) Ground truth.}
\end{figure}

We then qualitatively compare the SR quality of different light-weight models. For a more prominent comparison, we compare the $\times$4 SR results on the example image instead of $\times$2 SR, as shown in Fig. 2. From Fig. \ref{fig:CARN}, Fig. \ref{fig:IMDN} and Fig. \ref{fig:SwinIR}, we can see that SwinIR-light~\cite{liang2021swinir}  and all the CNN-based models result in blurry and distorted edges in ideal SR conditions. However, compared to Fig. \ref{fig:HR}, Fig. \ref{fig:ELAN} and Fig. \ref{fig:SIBSR} show that the transformer-based models can restore more accurate and clear structures and have the potential to recover clear and sharp edges in ideal SR conditions, demonstrating the effectiveness of introducing self-attention. It's worth mentioning that the visuals of the proposed model are almost comparable to ELAN-light~\cite{zhang2022efficient}, while parameters of the proposed model are fewest among the light-weight transformer-based SR models.
\subsection{Evaluations on Real-World Images with Unpredictable Blur Kernels}
\label{4.3}
For imitating the real-world images, we randomly generate various blur kernels to characterize different fuzzy conditions of real-world images caused by various unpredictable factors, and then, we evaluate the proposed framework on these blur kernel conditions. We assume four scenarios: severe aliasing, isotropic Gaussian, anisotropic Gaussian, and isotropic Gaussian, followed by \emph{bicubic subsampling}. PSNR and SSIM are used as evaluation metrics, which are calculated on the Y channel after converting RGB to YCbCr format.
\begin{itemize}
    \item [\textbullet] g$^d_{0.2}$ : isotropic Gaussian blur kernel with width $\lambda$ = 0.2 followed by \emph{direct} subsampling.
    \item [\textbullet] g$^d_{2.0}$ : isotropic Gaussian blur kernel with width $\lambda$ = 2.0 followed by \emph{direct} subsampling.
    \item [\textbullet] g$^d_{ani}$ : anisotropic Gaussian with widths $\lambda_1$ = 4.0 and $\lambda_2$ = 1.0 with $\Theta$ = -0.5 from
    \begin{equation}
        \sum = \left[\begin{array}{cc}
             cos(\Theta)&-sin(\Theta)  \\
             sin(\Theta)&cos(\Theta) 
        \end{array}\right] \left[\begin{array}{cc}
             \lambda_1&0  \\
             0&\lambda_2 
        \end{array}\right] \left[\begin{array}{cc}
             cos(\Theta)&sin(\Theta)  \\
             -sin(\Theta)&cos(\Theta) 
        \end{array}\right]
    \end{equation}
     followed by \emph{direct} subsampling.
    \item [\textbullet] g$^b_{1.3}$ : isotropic Gaussian blur kernel with width $\lambda$ = 1.3 followed by \emph{bicubic} subsampling.
\end{itemize}

\subsubsection{Quantitative comparison}
The results on various kernels are shown in Table \ref{tbl:blind_sr_tasks}. ZSSR~\cite{shocher2018zero} is a classical blind SR model, when it comes to the LR input images, it trains from scratch for each LR input image to recover the counterpart HR image. However, it requires thousands of gradient updates, which will increase the run time significantly. KernelGAN~\cite{bell2019blind} estimates the blur kernel, and then, generates the HR outputs with the help of ZSSR~\cite{shocher2018zero}. The performance drop caused by the inaccuracy of blur kernel estimation will be further amplified by ZSSR~\cite{shocher2018zero}. Our methods show better PSNR/SSIM than the regular blind SR methods, with the help of the proposed pre-trained transformer-based model, which enhances the representational power and accelerate the inference. Moreover, with the proposed fast coarse-to-fine training strategy, the number of gradient updates can be reduced to 10, from the origin 3000 in ZSSR~\cite{shocher2018zero} and KernelGAN~\cite{bell2019blind}, greatly reducing the running time.
\begin{table}[ht]
\centering
\caption{The average PSNR/SSIM results on various kernels with $\times$2 in real-world conditions. Best and second-best PSNR/SSIM indexes are marked in {\color{red}red} and {\color{blue}blue} colors, respectively.}
\resizebox{\linewidth}{!}{
\begin{tabular}{|c|c|c|c|c|c|c|c|}
\hline
Scale & Kernel & Dataset & ZSSR~\cite{shocher2018zero}         & KernelGAN~\cite{bell2019blind}    & ELAN~\cite{zhang2022efficient}         & SwinIR~\cite{liang2021swinir}       & S$^2$R (Ours)         \\ \hline
$\times$2    & g$^d_{0.2}$    & Set5    & 28.45/0.8592 & 21.11/0.5903 & \color{red}{28.51/0.8694} & 28.46/0.8686 & \color{blue}{28.47/0.8684} \\
$\times$2    & g$^d_{2.0}$    & Set5    & 29.16/0.8602 & 26.17/0.8074 & 29.17/0.8612 & \color{blue}{29.17/0.8613} & \color{red}{30.06/0.8890} \\
$\times$2    & g$^d_{ani}$    & Set5    & 28.41/0.8374 & 25.74/0.7802 & \color{blue}{28.44/0.8392} & 28.43/0.8390 & \color{red}{28.93/0.8583} \\
$\times$2    & g$^b_{1.3}$    & Set5    & $\color{blue}{31.59}$/0.8991 & 29.17/0.8811 & 31.54/0.9000 & 31.58/\color{blue}{0.9005} & \color{red}{33.37/0.9239}  \\ \hline
\end{tabular}
}
\label{tbl:blind_sr_tasks}
\end{table}

Furthermore, from Table \ref{tbl:blind_sr_onSISR}, we find that transformer-based models also show relatively good performance in real-world tasks, further proving the importance of introducing self-attention to extract features. At the g$^d_{0.2}$ condition, the blur kernel width is relatively small, resulting in a relatively incorrect blur kernel estimation, subsequently, amplified by the proposed coarse-to-fine training strategy, leading to a result that is a little bit worse than ELAN~\cite{zhang2022efficient}. Besides, the proposed framework shows better PSNR/SSIM than others.
\begin{table}[ht]
\centering
\caption{The average PSNR/SSIM results on ``bicubic" downsampling scenario with $\times$2 on benchmarks. Those with better PSNR/SSIM will be bolded. }
\begin{tabular}{|c|c|c|c|}
\hline
scale & Dataset  & ZSSR~\cite{shocher2018zero}         & S$^2$R (Ours)         \\ \hline
$\times$2    & Set5~\cite{bevilacqua2012low}     & 36.93/0.9554 & \textbf{38.15/0.9611} \\
$\times$2    & BSD100~\cite{martin2001database}   & 31.43/0.8901 & \textbf{32.30/0.9013} \\
$\times$2    & Urban100\cite{huang2015single} & 29.34/0.8941 & \textbf{32.61/0.9333} \\ \hline
\end{tabular}
\label{tbl:blind_sr_onSISR}
\end{table}
\subsubsection{Qualitative comparison}
\begin{figure}[h]
    \begin{minipage}[b]{0.99\textwidth}
    \centering
            \subfigure[ZSSR 34.95dB]{
            \includegraphics[width=0.3\textwidth]{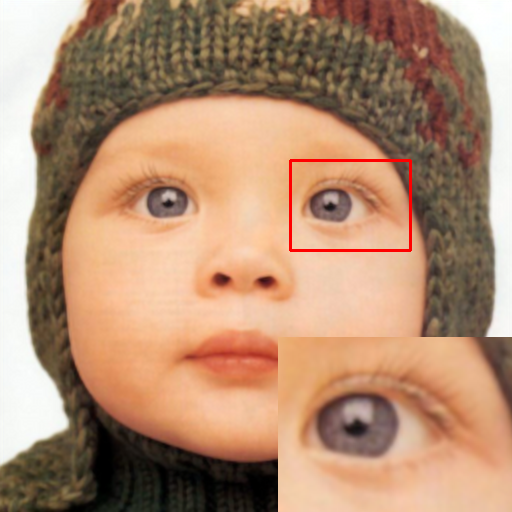}
            \label{fig:zssr}
        }
        \subfigure[KernelGAN 33.42dB]{
            \includegraphics[width=0.3\textwidth]{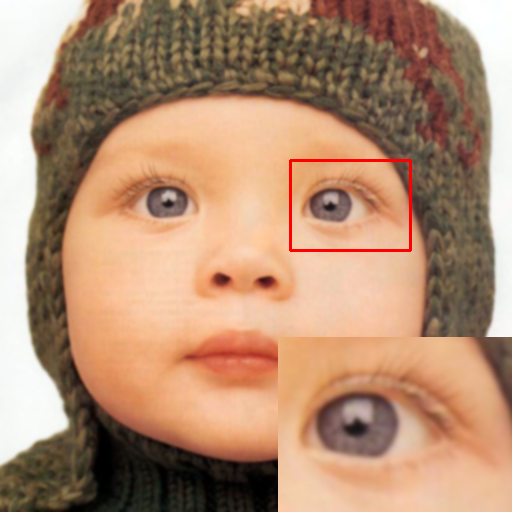}
            \label{fig:kernelgan}
        }
        \subfigure[ELAN-light 34.94dB]{
            \includegraphics[width=0.3\textwidth]{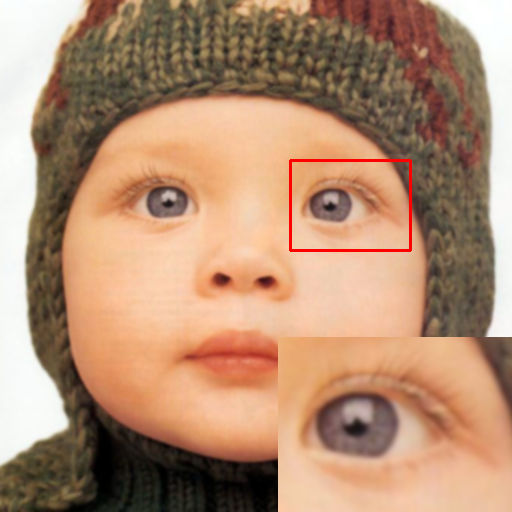}
            \label{fig:elan}
        }\\
        \subfigure[SwinIR-light 33.51dB]{
            \includegraphics[width=0.3\textwidth]{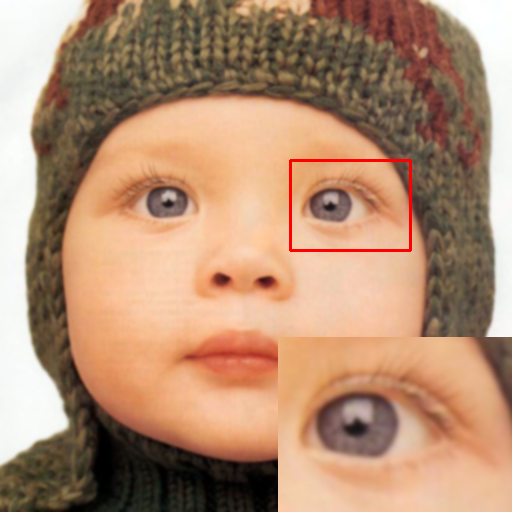}
            \label{fig:swinir}
        }
        \subfigure[S$^2$R (Ours) 37.09dB]{
            \includegraphics[width=0.3\textwidth]{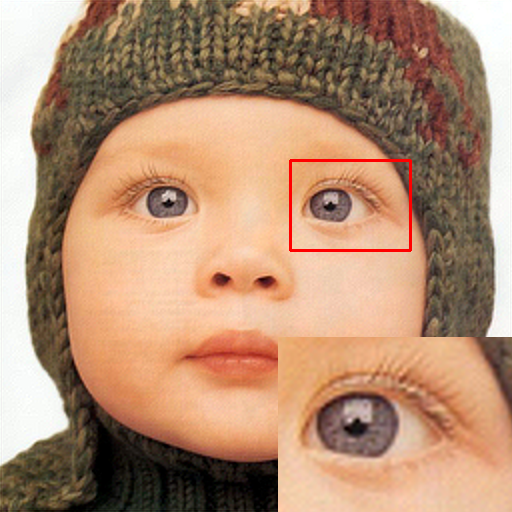}
            \label{fig:s2r}
        }
        \subfigure[GT]{
            \includegraphics[width=0.3\textwidth]{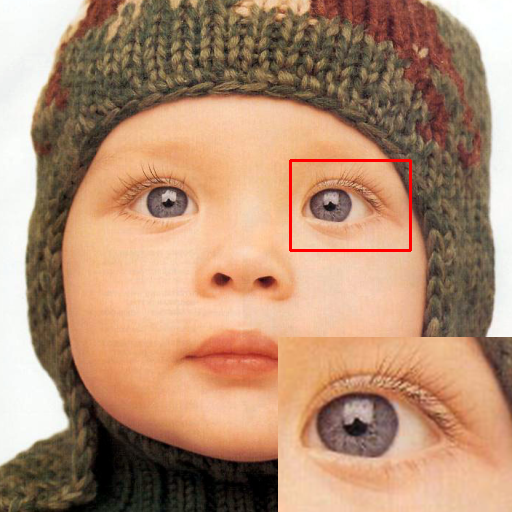}
            \label{fig:gt}
        }
    \end{minipage}
\label{baby_all}
\caption{Qualitative comparison of state-of-the-art blind SR methods for $\times$2 upscaling in real-world SR conditions with the kernel g$^b_{1.3}$, where the enlarged local patch of output image is placed in the lower right corner of the image. (a) The result of ZSSR~\cite{shocher2018zero}. (b) The result of KernelGAN~\cite{bell2019blind}. (c) The results of ELAN-light~\cite{zhang2022efficient}. (d)The result of SwinIR-light~\cite{liang2021swinir}.  (e) The result of ours. (f) Ground truth.}
\end{figure}

We then qualitatively compare the SR quality of different blind SR methods in real-world conditions with the kernel g$^b_{1.3}$. As shown in Fig. 3, the visual results outperform other methods. Our method can better mine the internal features of the image, obtaining sharper edges and higher contrast results. Compared with Fig. \ref{fig:zssr}, Fig. \ref{fig:kernelgan}, Fig. \ref{fig:elan} and Fig. \ref{fig:swinir}, the results of ours are far clearer than them, and closest to GT.
\subsection{Ablation Study}
In order to further verify the necessity of the modules in our network and our proposed training recipe, we conduct ablation experiments on the corresponding dataset.
\subsubsection{For Network Architecture}
\begin{itemize}
    \item [\textbullet] This setting only contains ELAB in the deep feature module, so it can be used to observe the effect of ELAB.
    \item[\textbullet] This setting contains ELAB and MBConv in the deep feature module, so it can be used to observe the effect of MBConv.
    \item[\textbullet] This setting is full model, containing ELAB, MBConv and Deform Conv, so it can be used to verify the effect of Deformable Conv.
\end{itemize}
The results are shown in Table \ref{ablation_arch}. Benefit from the efficient design of ELAB, compared with conventional SR models, such as SwinIR~\cite{liang2021swinir}, our model achieves comparable performance with fewer parameters. Moreover, based on the intuition that the bottlenecks contain all the necessary information, MBConv compensates for the performance drop caused by the reduced number of parameters. Finally, Deformable Conv enlarges the receptive field, further improving the task performance effectively.
\begin{table}[ht]
\centering
\caption{Ablation study on network design for S$^2$R transformer. Those with better PSNR/SSIM will be bolded.}
\resizebox{\linewidth}{!}{
\begin{tabular}{|c|c|ccc|c|c|c|c|c|c|}
\hline
\multirow{2}{*}{Scale} & \multirow{2}{*}{Model} & \multicolumn{3}{c|}{Different components}                             & \multirow{2}{*}{\#Params(K)} & Set5~\cite{bevilacqua2012low}         & Set14~\cite{zeyde2012single}        & BSD100~\cite{martin2001database}       & Urban100\cite{huang2015single}  & Manga109\cite{matsui2017sketch}  \\ \cline{3-5} \cline{7-11} 
                       &                        & \multicolumn{1}{c|}{ELAB} & \multicolumn{1}{c|}{MBConv} & Deform Conv &                              & PSNR/SSIM    & PSNR/SSIM    & PSNR/SSIM    & PSNR/SSIM & PSNR/SSIM \\ \hline
$\times$2                     & S$^2$R transformer                  & \multicolumn{1}{c|}{\checkmark}    & \multicolumn{1}{c|}{}       &             & 523.45                       & 38.12/0.9609 & 33.80/0.9203 & 32.27/0.9009 & 32.45/0.9325 & 38.85/0.9777          \\
$\times$2                     & S$^2$R transformer                  & \multicolumn{1}{c|}{\checkmark}    & \multicolumn{1}{c|}{\checkmark}      &             & 562.94                       &  38.14/0.9610  &  33.83/0.9205 & 32.28/0.9010 & 32.50/0.9330 & 38.95/0.9780 \\
$\times$2                     & S$^2$R transformer                  & \multicolumn{1}{c|}{\checkmark}    & \multicolumn{1}{c|}{\checkmark}      & \multicolumn{1}{c|}{\checkmark}           & 578                       & \textbf{38.15/0.9611} & \textbf{33.88/0.9206} & \textbf{32.30/0.9013} &  \textbf{32.61/0.9333} & \textbf{39.01/0.9780}   \\ \hline      
\end{tabular}
}

\label{ablation_arch}
\end{table} 
 
\subsubsection{For Training Strategy}
\begin{itemize}
    \item [\textbullet] In this setting, we directly use the proposed model to deal with the real-world tasks without the proposed fast training strategy.
    \item [\textbullet] In this setting, the proposed model is pre-trained, and then fine-level transfer learning is proposed to adapt to the real-world blur kernel. This setting is used to verify the effect of the proposed coarse-to-fine training strategy.
\end{itemize}
The key point of our training recipe is to fine-tune on several other blur kernels after pre-training our model. Our purpose is to broaden the scalability of our model and reduce the runtime simultaneously. With the help of self-similarity, transformer-based models benefit the performance on blind SR tasks. Moreover, the per-trained model avoids up to 3000 backpropagation gradient updates, reducing the runtime significantly.
\begin{table}[ht]
\centering
\caption{Ablation study on training strategy for S$^2$R transformer. Those with better PSNR/SSIM will be bolded.}
\resizebox{\linewidth}{!}{
\begin{tabular}{|c|c|l|c|c|c|c|}
\hline
\multirow{2}{*}{Scale} & \multirow{2}{*}{Model} & \multicolumn{1}{c|}{\multirow{2}{*}{\begin{tabular}[c]{@{}c@{}}Trainging\\ Strategy\end{tabular}}} & Set5(g$^d_{0.2}$)    & Set5(g$^d_{2.0}$)     & Set5(g$^d_{ani}$)    & Set5(g$^b_{1.3}$)     \\ \cline{4-7} 
                       &                        & \multicolumn{1}{c|}{}                                                                              & PSNR/SSIM    & PSNR/SSIM    & PSNR/SSIM    & PSNR/SSIM    \\ \hline
$\times$2                     & S$^2$R transformer                 &                                                                                                    & 28.36/0.8664 & 29.17/0.8614 & 28.43/0.8391 & 31.58/0.9007 \\
$\times$2                     & S$^2$R transformer                 & \multicolumn{1}{c|}{\checkmark}                                                                                                  & \textbf{28.47/0.8684} & \textbf{30.06/0.8890} & \textbf{28.93/0.8583} & \textbf{33.37/0.9239} \\ \hline
\end{tabular}
}
\end{table}
\section{Conclusion}
In this paper, we proposed a novel framework S$^2$R, which consists of a light-weight transformer-based model and a novel coarse-to-fine training strategy, achieving excellent performance and good visual results in both SISR and blind SR tasks. 
On one hand, some light-weight blocks are utilized in the proposed model to reduce the number of parameters further while maintaining a great performance. On the other hand, the proposed training strategy broadens the scalability of the proposed model in real-world situations. From our extensive experiments, regardless of whether in real-world applications or in ideal conditions, the proposed model outperforms previous methods in terms of visual results with the lowest number of parameters 578K. Moreover, the proposed framework accelerates the transfer-learning process in real-world situations by extremely reducing the number of backpropagation gradient updates, Compared to ZSSR~\cite{shocher2018zero}, the reduction is as high as 300 times.


\end{document}